\documentclass[11pt]{article}
\usepackage[utf8]{inputenc}
\usepackage{amsmath}
\usepackage{graphicx}
\usepackage{booktabs}
\usepackage[
    colorlinks=true,
    linkcolor=blue,
    citecolor=blue,
    urlcolor=blue,
    pdfstartview=FitH,
    linktoc=all
]{hyperref}
\usepackage{listings}
\usepackage{geometry}
\title{Agent-Agnostic Evaluation of SQL Accuracy in\\Production Text-to-SQL Systems}

\author{
    Taslim Jamal Arif \\
    \texttt{tjarif743@gmail.com}
    \and 
    Kuldeep Singh \\
    \texttt{ksksingh022@gmail.com}
}

\date{April 2026}

\begin{document}

\maketitle

\begin{abstract}
Text-to-SQL (T2SQL) evaluation in production environments poses fundamental challenges that existing benchmarks do not address. Current evaluation methodologies  whether rule-based SQL matching or schema-dependent semantic parsers  assume access to ground-truth queries and structured database schema, constraints that are rarely satisfied in real-world deployments. This disconnect leaves production T2SQL agents largely unevaluated beyond developer-time testing, creating silent quality degradation with no feedback mechanism for continuous improvement. We present STEF (Schema-agnostic Text-to-SQL Evaluation Framework), a production-native evaluation system that operates exclusively on natural language inputs  the user question, an enriched reformulation, and the generated SQL  without requiring database schema or reference queries. STEF extracts semantic specifications from both natural language and SQL representations, performs normalized feature alignment, and produces an interpretable 0 to 100 accuracy score via a composite metric that encompasses filter alignment, semantic verdict, and confidence of the evaluator. Key contributions include: enriched question quality validation as a first-class evaluation signal, configurable application-specific rule injection via prompt templating, and production-robust normalization handling GROUP BY tolerance, ORDER BY defaults, and LIMIT heuristics. Empirical results demonstrate that STEF enables continuous production monitoring and agent improvement feedback loops without schema dependency, making structured query evaluation viable at scale for the first time.
\end{abstract}

\section{Introduction}

Text-to-SQL (T2SQL) systems represent a critical interface layer between natural language 
and relational databases, enabling non-technical stakeholders to extract structured insights 
through conversational queries. As enterprise adoption of large language model (LLM)-powered 
data assistants accelerates, the ability to accurately evaluate T2SQL generation quality in 
live production settings has become a foundational requirement for responsible deployment 
and continuous system improvement.

Despite significant advances in T2SQL research  driven by benchmark datasets such as 
Spider~\cite{yu2018spider}, WikiSQL~\cite{wikisql}, and BIRD~\cite{bird}  a fundamental disconnect 
persists between academic evaluation paradigms and the operational realities of production 
systems. Existing evaluation methodologies broadly fall into two categories: \textit{rule-based 
exact or partial match} approaches~\cite{yu2018spider}, which compare generated SQL 
against a reference query using string or structural canonicalization, and 
\textit{execution-based accuracy} (EX) methods~\cite{execution_accuracy}, which execute 
both the generated and reference SQL against a live or snapshot database and compare result 
sets. Both paradigms share a critical assumption: access to ground-truth SQL queries and, 
in most execution-based settings, a queryable database schema.

In real-world production deployments, neither of these assumptions holds reliably. Reference 
SQL queries are rarely available at inference time  production T2SQL agents are invoked 
continuously by end users whose questions have no pre-annotated gold standard. Database 
schemas may be proprietary, multi-tenant, or subject to frequent schema evolution, making 
static schema dependency a significant operational liability.

These gaps leave production T2SQL systems operating in an evaluation vacuum: developers 
can test agents at development time using curated benchmarks, but once deployed, there 
exists no systematic mechanism to detect accuracy degradation, compare agent quality across 
model versions, or identify categories of queries that consistently underperform. This 
silent quality erosion represents a significant risk in enterprise settings where data 
decisions are made directly from T2SQL outputs.

To address these challenges, we present \textbf{STEF} (\textbf{S}chema-agnostic 
\textbf{T}ext-to-SQL \textbf{E}valuation \textbf{F}ramework), a production-native 
evaluation system designed to operate exclusively on natural language signals  the 
original user question, the agent-generated enriched reformulation, and the produced 
SQL query  without requiring access to database schema, reference queries, or query 
execution infrastructure. STEF is architected around three core technical contributions:

\begin{enumerate}

    \item \textbf{Normalized Semantic Feature Extraction and Alignment.} 
    STEF extracts structured semantic specifications from both the natural language 
    question and the generated SQL independently, then performs normalized feature-level 
    alignment across projections, aggregations, filter conditions, and grouping semantics. 
    By decomposing evaluation into orthogonal semantic dimensions rather than treating 
    the SQL as a monolithic string, STEF produces fine-grained diagnostic signals that 
    identify precisely \textit{which} semantic component of a query is misaligned, 
    enabling targeted agent improvement.

    \item \textbf{Configurable Application-Specific Rule Injection.} 
    Different enterprise deployments exhibit distinct SQL conventions: column naming 
    standards, benign filter defaults, domain-specific aggregation patterns, and 
    acceptable query structural variations that should not be penalized. STEF supports 
    runtime injection of application rules via prompt templating, allowing evaluation 
    behavior to be customized per deployment without code changes or model retraining.
\end{enumerate}

Evaluation is expressed as a composite interpretable score on a 0--100 scale, derived 
from filter alignment status, semantic verdict, and a calibrated confidence multiplier 
that accounts for LLM evaluator uncertainty. Production normalization rules handle 
common patterns  including \texttt{GROUP BY} inference, \texttt{ORDER BY} defaults, 
and \texttt{LIMIT} heuristics  that would otherwise generate false negatives against 
reasonable, semantically correct SQL.

The remainder of this paper is organized as follows. Section~\ref{sec:related} surveys 
related work in T2SQL evaluation and LLM-as-judge frameworks. Section~\ref{sec:method} 
details the STEF evaluation pipeline, enrichment validation, and feature extraction 
methodology. Section~\ref{sec:scoring} presents the composite scoring formulation. 
Section~\ref{sec:normalization} describes production normalization rules. 
Section~\ref{sec:impl} covers implementation and prompt engineering. 
Section~\ref{sec:results} presents empirical results. Section~\ref{sec:discussion} 
discusses strengths, limitations, and future directions, and Section~\ref{sec:conclusion} 
concludes.

\section{Related Work}
\label{sec:related}

The evaluation of Text-to-SQL systems has been an active area of research, evolving 
alongside the rapid progression of semantic parsing and large language model capabilities. 
We organize the related literature across four dimensions: classical string-based evaluation, 
execution-based metrics, semantic and structural alignment approaches, and LLM-as-judge 
frameworks. We additionally review production-oriented evaluation efforts and highlight 
the gaps that motivate our work.

\subsection{String-Based and Structural Matching}

Early T2SQL evaluation relied on \textit{exact match} (EM) metrics, which compare the 
generated SQL string directly against a reference query after normalization. While 
straightforward to compute, exact match is well-known to penalize semantically equivalent 
queries that differ in token ordering, alias naming, or subquery structure~\cite{yu2018spider}. 
Yu et al.~\cite{yu2018spider} introduced the Spider benchmark alongside a 
\textit{component-level exact match} metric that independently evaluates SQL clauses  
\texttt{SELECT}, \texttt{WHERE}, \texttt{GROUP BY}, \texttt{ORDER BY}, and 
\texttt{HAVING}  reducing sensitivity to superficial syntactic variation. Despite this 
improvement, component-level EM still requires access to gold-standard SQL and is 
brittle to paraphrastic variation in column references and filter expressions. Subsequent 
works such as SParC~\cite{sparc} and CoSQL~\cite{cosql} extended the Spider paradigm to 
multi-turn and conversational settings, but retained the same reference-query dependency 
that limits applicability in schema-free production environments.

\subsection{Execution-Based Accuracy}

Execution Accuracy (EX)~\cite{execution_accuracy} metrics address the surface-form sensitivity of string matching 
by instead comparing the \textit{result sets} produced by executing the generated and 
reference SQL against a shared database. This approach 
correctly treats semantically equivalent but syntactically distinct queries as equivalent, 
and has become the dominant evaluation paradigm in benchmarks such as 
Spider~\cite{yu2018spider}, BIRD~\cite{bird}, and WikiSQL~\cite{wikisql}. However, execution 
accuracy introduces its own limitations in production settings. First, it requires access 
to a live or snapshot database against which both queries can be executed  an assumption 
that fails in multi-tenant enterprise deployments, federated data environments, or when 
schemas undergo continuous evolution. Second, execution equality does not imply semantic 
equivalence: two queries may return identical result sets on a particular database instance 
while expressing fundamentally different intents, leading to false positives~\cite{bird}. 
Third, no reference SQL is available for user queries arriving in production, making 
execution comparison operationally infeasible without significant annotation overhead.

\subsection{Semantic Parsing and Structural Alignment}

A complementary line of research focuses on structured intermediate representations 
that abstract away syntactic variation while preserving semantic content. IRNet~\cite{irnet} 
introduced SemQL, an intermediate SQL-like representation designed to facilitate 
schema-linking and reduce sensitivity to database-specific column naming. RATSQL~\cite{ratsql} 
and subsequent graph-neural-network-based parsers improved schema linking by modeling 
relationships between question tokens and schema elements as a graph, but this approach 
inherently requires schema access as a structural input. More recently, the BIRD 
benchmark~\cite{bird} highlighted the importance of \textit{efficiency} and 
\textit{value matching} as additional evaluation dimensions beyond structural correctness, 
noting that production queries are subject to real-world database efficiency constraints 
not captured by academic evaluation metrics. These contributions advance structural 
alignment but remain tightly coupled to schema availability, limiting their applicability 
to our target setting.

\subsection{LLM-as-Judge Evaluation Frameworks}

The emergence of capable large language models has motivated a growing body of work on 
\textit{LLM-as-judge} evaluation, where a separate LLM is prompted to assess the quality 
of a primary model's output~\cite{llm_judge_variance}. Zheng et al.~\cite{mtbench} demonstrated 
that GPT-4 as a judge exhibits strong correlation with human preference judgments across 
a variety of NLP tasks, establishing the viability of the paradigm. In the T2SQL domain, 
Snowflake's Cortex Analyst~\cite{snowflake_cortex} employs an LLM-based evaluation layer 
to assess semantic alignment between natural language questions and generated SQL without 
full schema dependency. However, Cortex Analyst is designed as an integrated component 
of Snowflake's proprietary data cloud infrastructure and does not generalize to 
agent-agnostic or cross-platform deployments. IBM Research~\cite{ibm2025} has similarly 
explored production SQL evaluation frameworks leveraging LLM reasoning, but without 
addressing enriched question validation or configurable application-specific rule injection.

A key challenge in LLM-as-judge approaches is \textit{evaluator variance}: the same 
query pair may receive inconsistent judgments across LLM calls due to temperature 
sensitivity and prompt formulation effects~\cite{llm_judge_variance}. STEF mitigates 
this through structured output elicitation with explicit confidence scoring, followed by 
a confidence-weighted multiplier in the composite score, transforming evaluator 
uncertainty from a noise source into an interpretable signal.

\section{Methodology}
\label{sec:method}

STEF operates as a multi-stage evaluation pipeline that transforms raw natural language 
and SQL inputs into a structured semantic comparison, producing an interpretable accuracy 
score without requiring database schema or reference queries. This section describes each stage of the 
pipeline in detail: input specification, enriched question validation, semantic feature 
extraction, normalized specification alignment, and application-specific rule injection.

\subsection{System Inputs}
\label{sec:inputs}

The evaluation pipeline accepts four inputs for each query instance:

\begin{itemize}
    \item $Q_u$: The \textbf{original user question} as submitted to the T2SQL agent, 
    representing the unmodified natural language intent of the end user.
    
    \item $Q_e$: The \textbf{enriched question} produced by the T2SQL agent's preprocessing 
    pipeline.
    
    \item $\mathcal{S}$: The \textbf{generated SQL query} produced by the T2SQL agent 
    in response to the enriched question.
    
    \item $\mathcal{R}_{app}$: An \textbf{application-specific rule configuration}, 
    provided as a structured JSON object injected at prompt construction time, encoding 
    deployment-specific column mappings, benign filter defaults, and evaluation tolerance 
    parameters.
\end{itemize}

Critically, no database schema, table metadata, or reference SQL is required at any 
stage. The framework is designed to be fully self-contained with respect to these four 
inputs, making it deployable across heterogeneous database technologies and T2SQL agent 
architectures without modification.

\subsection{Stage 2: Semantic Feature Extraction}
\label{sec:feature_extraction}

STEF extracts structured semantic specifications 
independently from two sources: the natural language representation (combining $Q_u$ 
and $Q_e$) and the generated SQL $\mathcal{S}$. This dual extraction strategy enables 
feature-level alignment that is more robust to surface-form variation than direct SQL 
string comparison, and more diagnostically informative than coarse verdict-only 
judgments.

\subsubsection{Question-Side Specification ($\mathcal{S}_q$)}

The question-side specification captures the analytical intent expressed in natural 
language, decomposed into four semantic components:

\begin{align}
    \mathcal{S}_q.\mathtt{outputs} &= \{c_1, c_2, \ldots, c_n\} 
    \label{eq:sq_outputs} \\
    \mathcal{S}_q.\mathtt{aggregations} &= \{(\mathtt{func}_i,\; \mathtt{col}_i,\; 
    \mathtt{distinct}_i)\}_{i=1}^{m} 
    \label{eq:sq_aggs} \\
    \mathcal{S}_q.\mathtt{filters} &= \{(\mathtt{lhs}_j,\; \mathtt{op}_j,\; 
    \mathtt{rhs}_j)\}_{j=1}^{k} 
    \label{eq:sq_filters} \\
    \mathcal{S}_q.\mathtt{group\_by} &= \{g_1, g_2, \ldots, g_p\}
    \label{eq:sq_groupby}
\end{align}

\noindent where $\mathcal{S}_q.\mathtt{outputs}$ enumerates the columns or derived 
metrics that the user question explicitly or implicitly requests; 
$\mathcal{S}_q.\mathtt{aggregations}$ captures aggregation intent expressed in natural 
language (e.g., ``total spend'', ``average revenue'', ``distinct customer count''); 
$\mathcal{S}_q.\mathtt{filters}$ enumerates filter conditions implied by the question 
phrasing (e.g., ``for the year 2023'', ``in the APAC region'', ``where product category 
is Electronics''); and $\mathcal{S}_q.\mathtt{group\_by}$ captures any grouping 
dimensions explicitly or implicitly requested (e.g., ``broken down by country'', 
``per quarter'').

\subsubsection{SQL-Side Specification ($\mathcal{S}_{sql}$)}

An analogous specification is extracted from the generated SQL query through structured 
parsing of the \texttt{SELECT}, \texttt{WHERE}, \texttt{GROUP BY}, \texttt{HAVING}, 
\texttt{ORDER BY}, and \texttt{LIMIT} clauses:

\begin{align}
    \mathcal{S}_{sql}.\mathtt{projections} &= \{p_1, p_2, \ldots, p_n\} 
    \label{eq:ssql_proj} \\
    \mathcal{S}_{sql}.\mathtt{aggregations} &= \{(\mathtt{func}_i,\; \mathtt{col}_i,\; 
    \mathtt{distinct}_i)\}_{i=1}^{m} 
    \label{eq:ssql_aggs} \\
    \mathcal{S}_{sql}.\mathtt{filters} &= \{(\mathtt{lhs}_j,\; \mathtt{op}_j,\; 
    \mathtt{rhs}_j)\}_{j=1}^{k} 
    \label{eq:ssql_filters} \\
    \mathcal{S}_{sql}.\mathtt{group\_by} &= \{g_1, g_2, \ldots, g_p\} 
    \label{eq:ssql_groupby} \\
    \mathcal{S}_{sql}.\mathtt{order\_by} &= \{(o_l,\; \mathtt{dir}_l)\}_{l=1}^{r} 
    \label{eq:ssql_orderby} \\
    \mathcal{S}_{sql}.\mathtt{limit} &= \lambda
    \label{eq:ssql_limit}
\end{align}

SQL parsing is performed by the LLM evaluator, which is explicitly instructed to assume 
that all referenced table and column names exist in the underlying schema. This 
assumption is necessary to enable schema-free operation and prevents the evaluator from 
generating false negatives based on unverifiable schema validity claims.

\subsubsection{Normalization During Extraction}

To enable robust alignment between $\mathcal{S}_q$ and $\mathcal{S}_{sql}$, both 
specifications undergo normalization during extraction. Normalization handles the 
following common production patterns:

\begin{itemize}
    \item \textbf{Semantic alias resolution:} Column aliases in SQL (e.g., 
    \texttt{SUM(spend) AS total\_spend}) are resolved to their underlying semantic 
    meaning for alignment with natural language aggregation expressions.
    
    \item \textbf{Filter value normalization:} String filter values are normalized 
    for case and whitespace, and pattern-matching operators (\texttt{ILIKE}, 
    \texttt{LIKE}) are treated as semantically equivalent to equality filters for 
    alignment purposes when the pattern is non-wildcarded.
    
    \item \textbf{Implicit aggregation inference:} Natural language expressions such 
    as ``total'', ``sum of'', ``average'', and ``count of'' are mapped to their 
    corresponding SQL aggregation functions (\texttt{SUM}, \texttt{AVG}, \texttt{COUNT}) 
    during question-side extraction, enabling direct comparison with SQL-side 
    aggregation specifications.
    
    \item \textbf{Application rule substitution:} Column name mappings specified in 
    $\mathcal{R}_{app}$ are applied during extraction so that domain-specific aliases 
    (e.g., \texttt{region} $\rightarrow$ \texttt{RegionName}) are resolved to their 
    canonical form before alignment is performed.
\end{itemize}

\subsection{Stage 3: Specification Alignment and Filter Status Classification}
\label{sec:alignment}

With both specifications extracted and normalized, STEF performs pairwise alignment 
across each semantic dimension to produce a structured comparison record. The most 
diagnostically significant alignment dimension is \textit{filter alignment}, which 
determines whether the filter conditions implied by the user question are correctly 
reflected in the generated SQL.

Filter alignment produces one of four classification outcomes:

\begin{itemize}
    \item \textbf{Fully Applied:} All filters in $\mathcal{S}_q.\mathtt{filters}$ 
    are present in $\mathcal{S}_{sql}.\mathtt{filters}$ with matching left-hand side, 
    operator, and right-hand side after normalization, and no extraneous filters are 
    present that contradict the question intent.
    
    \item \textbf{Fully Applied with Extras:} All required filters are correctly 
    present, but the SQL additionally contains filter conditions not implied by the 
    question. Extras are assessed against $\mathcal{R}_{app}.\mathtt{benign\_filters}$ 
    to determine whether they represent acceptable application defaults or potentially 
    unintended restrictions.
    
    \item \textbf{Partially Applied:} A subset of required filters is correctly 
    present, with one or more required filters absent or incorrectly specified. 
    Partial application is further annotated with the set of missing and mismatched 
    filters for diagnostic purposes.
    
    \item \textbf{Not Applied:} No required filter conditions are reflected in the 
    generated SQL, indicating a complete failure to translate filter intent.
\end{itemize}

Alignment is additionally performed across projection, aggregation, and grouping 
dimensions, with each producing a binary match signal that contributes to the overall 
semantic verdict produced in Stage 4.

\subsection{Stage 4: LLM-Based Semantic Verdict}
\label{sec:verdict}

Following structured feature alignment, STEF invokes an LLM evaluator to produce a 
holistic semantic verdict that captures alignment dimensions not amenable to 
deterministic rule-based assessment  including implicit metric equivalences, 
paraphrastic column references, and complex filter logic involving date arithmetic 
or nested conditions.

The LLM evaluator is provided with the structured comparison record produced in 
Stage 3, the original inputs $Q_u$, $Q_e$, and $\mathcal{S}$, and the application 
rule configuration $\mathcal{R}_{app}$. It is explicitly instructed to:

\begin{enumerate}
    \item Assume all referenced table and column names are valid.
    \item Apply the column mappings and benign filter definitions in $\mathcal{R}_{app}$ 
    before assessing correctness.
    \item Produce a verdict from a fixed four-point ordinal scale: \textit{Correct}, 
    \textit{Likely Correct}, \textit{Potentially Incorrect}, or \textit{Incorrect}.
    \item Provide an explicit confidence score $\gamma \in [0, 1]$ reflecting the 
    evaluator's certainty in the verdict, grounded in the clarity of the feature 
    alignment evidence.
    \item Return all outputs as a structured JSON object to enable deterministic 
    downstream scoring.
\end{enumerate}

The structured output format enforces consistency across evaluator invocations and 
enables the confidence score to be used as a calibrated weighting signal in the 
composite scoring function described in Section~\ref{sec:scoring}.

\subsection{Stage 5: Application-Specific Rule Injection}
\label{sec:rule_injection}

A key design principle of STEF is that production SQL evaluation cannot be divorced 
from deployment-specific context. Different enterprise applications impose different 
conventions on SQL structure, column naming, and acceptable query patterns. Hardcoding 
these conventions into the evaluation framework would sacrifice generalizability; 
ignoring them would generate systematic false negatives for correct, convention-compliant 
SQL.

STEF resolves this tension through a \textit{configurable rule injection} mechanism. 
Application rules are specified as a structured JSON configuration object 
$\mathcal{R}_{app}$ and injected directly into the evaluation prompt template at 
construction time. The rule schema supports three categories of configuration:

\begin{itemize}
    \item \textbf{Column Mappings} (\texttt{column\_mappings}): A dictionary mapping 
    natural language column references to their physical database column names 
    (e.g., \texttt{\{"region": "RegionName", "spend": "TotalSpendUSD"\}}). These 
    mappings are applied during both feature extraction and alignment to ensure that 
    semantically equivalent references are not treated as mismatches due to naming 
    conventions.

    \item \textbf{Benign Filters} (\texttt{benign\_filters}): A list of filter 
    conditions that are acceptable application defaults even when not explicitly 
    requested in the user question (e.g., \texttt{["status = 'Active'", 
    "is\_deleted = 0"]}). SQL containing these filters is not penalized under the 
    \textit{Fully Applied with Extras} classification.

    \item \textbf{Ignored Filters} (\texttt{ignore\_filters}): A list of filter 
    dimensions that should be excluded from alignment assessment entirely, typically 
    corresponding to multi-tenancy or access control filters injected by the 
    application layer rather than the T2SQL agent (e.g., \texttt{["portfolio"]}).
\end{itemize}

The rule injection mechanism is entirely prompt-driven, requiring no code changes to 
the STEF evaluation pipeline when application configurations change. This design 
supports zero-downtime reconfiguration and enables the same evaluation infrastructure 
to serve multiple independent T2SQL applications simultaneously with different rule 
profiles.

\begin{lstlisting}[language=Python, caption={Application Rule Configuration Schema}, 
label={lst:app_rules}]
APP_RULES = {
    "column_mappings": {
        "region":   "RegionName",
        "spend":    "TotalSpendUSD",
        "customer": "CustomerAccountID"
    },
    "benign_filters": [
        "status = 'Active'",
        "is_deleted = 0"
    ],
    "ignore_filters": [
        "portfolio",
        "tenant_id"
    ]
}
\end{lstlisting}

\section{Composite Scoring}
\label{sec:scoring}

STEF produces a single interpretable accuracy score on a 0--100 scale for each evaluated 
query instance. This score is designed to satisfy three properties that are essential for 
production monitoring utility: \textit{interpretability}, such that the score can be 
decomposed into its constituent signals for diagnostic purposes; \textit{calibration}, 
such that evaluator uncertainty is reflected in the score magnitude rather than silently 
absorbed; and \textit{sensitivity}, such that the score differentiates meaningfully 
between qualitatively distinct correctness levels rather than collapsing them into a 
binary outcome.

The composite score is derived from three orthogonal components: a \textit{filter 
alignment score}, a \textit{semantic verdict score}, and a \textit{confidence 
multiplier}. Each component captures a distinct and non-redundant dimension of 
evaluation quality, and their combination into a single scalar is designed to preserve 
the diagnostic information carried by each dimension.

\subsection{Score Formulation}
\label{sec:score_formulation}

The composite accuracy score $\Phi$ is defined as:

\begin{equation}
    \Phi = \frac{\mathcal{B}}{10} \times 100 \times \Gamma
    \label{eq:composite_score}
\end{equation}

\noindent where $\mathcal{B}$ denotes the \textit{base score}, computed as the sum of 
the filter alignment score, the semantic verdict score, and an optional leniency boost, 
and $\Gamma$ denotes the \textit{confidence multiplier} derived from the evaluator's 
reported confidence. The base score $\mathcal{B}$ is formally defined as:

\begin{equation}
    \mathcal{B} = \sigma_{\mathtt{filters}} + \sigma_{\mathtt{verdict}} + 
    \delta_{\mathtt{lenient}}
    \label{eq:base_score}
\end{equation}

\noindent where $\sigma_{\mathtt{filters}} \in \{0, 3, 4, 5\}$ is the filter alignment 
score, $\sigma_{\mathtt{verdict}} \in \{0, 2, 3, 5\}$ is the semantic verdict score, 
and $\delta_{\mathtt{lenient}} \in \{0, 1\}$ is a leniency boost applied in cases where 
normalization rules determine that a borderline filter classification is attributable to 
a benign production SQL pattern rather than a genuine semantic error. The maximum 
achievable base score without the leniency boost is $\mathcal{B}_{\max} = 10$, ensuring 
that Equation~\ref{eq:composite_score} maps the base score to the $[0, 100]$ range 
prior to confidence adjustment.

\subsection{Filter Alignment Score}
\label{sec:filter_score}

The filter alignment score $\sigma_{\mathtt{filters}}$ is assigned based on the filter 
status classification produced in Stage 3 of the evaluation pipeline 
(Section~\ref{sec:alignment}). The scoring assignment reflects the relative severity 
of each filter alignment outcome as a proxy for semantic correctness:

\begin{table}[h]
\centering
\caption{Filter Alignment Status and Corresponding Scores}
\label{tab:filter_scores}
\begin{tabular}{lcp{8cm}}
\toprule
\textbf{Filter Status} & \textbf{Score ($\sigma_{\mathtt{filters}}$)} & 
\textbf{Interpretation} \\
\midrule
\texttt{fully\_applied} & 5 & All required filters are correctly and completely 
reflected in the generated SQL with no unintended extras. \\[6pt]
\texttt{fully\_applied\_with\_extras} & 4 & All required filters are present, but 
the SQL includes additional filter conditions beyond those implied by the user 
question. Extras are assessed against the benign filter configuration. \\[6pt]
\texttt{partially\_applied} & 3 & A subset of required filters is correctly present, 
but one or more required conditions are absent or incorrectly specified. \\[6pt]
\texttt{not\_applied} & 0 & No required filter conditions are reflected in the 
generated SQL, indicating a complete failure to translate filter intent from the 
natural language question. \\
\bottomrule
\end{tabular}
\end{table}

The scoring gap between \texttt{partially\_applied} (3) and \texttt{not\_applied} (0) 
is intentionally large, reflecting the qualitative difference between partial 
filter coverage  which preserves some semantic constraint  and total filter 
omission, which produces a structurally unconstrained query that may return 
arbitrarily incorrect result sets in production.

The scoring gap between \texttt{fully\_applied} (5) and 
\texttt{fully\_applied\_with\_extras} (4) is deliberately narrow, reflecting the 
observation that extra filters in production SQL are frequently benign application 
defaults (e.g., soft-delete guards, active-status constraints) rather than semantic 
errors. When extra filters are confirmed benign via $\mathcal{R}_{app}$, the 
leniency boost $\delta_{\mathtt{lenient}} = 1$ may be applied to recover the full 
base score, as described in Section~\ref{sec:leniency}.

\subsection{Semantic Verdict Score}
\label{sec:verdict_score}

The semantic verdict score $\sigma_{\mathtt{verdict}}$ is assigned based on the 
four-point ordinal verdict produced by the LLM evaluator in Stage 4 
(Section~\ref{sec:verdict}). The verdict scale is designed to capture a continuum of 
correctness confidence rather than a binary correct/incorrect judgment, enabling the 
scoring function to preserve the evaluator's expressed uncertainty:

\begin{table}[h]
\centering
\caption{Semantic Verdict and Corresponding Scores}
\label{tab:verdict_scores}
\begin{tabular}{lcp{8cm}}
\toprule
\textbf{Verdict} & \textbf{Score ($\sigma_{\mathtt{verdict}}$)} & 
\textbf{Interpretation} \\
\midrule
\texttt{Correct} & 5 & The generated SQL fully and unambiguously satisfies the 
semantic intent of the user question, with all projections, aggregations, filters, 
and groupings correctly specified. \\[6pt]
\texttt{Likely Correct} & 3 & The generated SQL is semantically well-aligned with 
the user intent, with minor ambiguities or non-critical structural differences that 
do not materially affect the result. \\[6pt]
\texttt{Potentially Incorrect} & 2 & The generated SQL shows partial semantic 
alignment but contains one or more conditions, projections, or aggregations that 
may produce incorrect results in at least some database states. \\[6pt]
\texttt{Incorrect} & 0 & The generated SQL fundamentally misaligns with the user 
intent, either through incorrect metric selection, wrong filter logic, or structural 
errors that will produce systematically incorrect results. \\
\bottomrule
\end{tabular}
\end{table}

The verdict score intentionally assigns a non-trivial score (2) to 
\texttt{Potentially Incorrect} rather than treating it equivalently to 
\texttt{Incorrect}. This design choice reflects the production reality that partial 
correctness  where a query returns a result set that is partially aligned with user 
intent  is meaningfully distinct from complete incorrectness, and should be 
represented as such in continuous monitoring dashboards. It also ensures that 
borderline cases assessed with low evaluator confidence are naturally penalized 
further through the confidence multiplier rather than through an artificially 
inflated verdict score.

\subsection{Confidence Multiplier}
\label{sec:confidence_multiplier}

The confidence multiplier $\Gamma$ is a piecewise function of the evaluator's 
reported confidence score $\gamma \in [0, 1]$, defined as:

\begin{equation}
    \Gamma(\gamma) = 
    \begin{cases}
        1.0 & \text{if } \gamma \geq 0.85 \\
        0.8 & \text{if } 0.65 \leq \gamma < 0.85 \\
        0.5 & \text{if } \gamma < 0.65
    \end{cases}
    \label{eq:conf_multiplier}
\end{equation}

The confidence multiplier serves a dual function. First, it transforms evaluator 
uncertainty from a noise source into an explicit scoring signal: queries where the 
LLM evaluator expresses low confidence are penalized in proportion to that 
uncertainty, ensuring that high scores are only achievable when the evaluator is 
strongly confident in its assessment. Second, it provides a natural segmentation 
of evaluated queries into three reliability tiers  \textit{high confidence} 
($\gamma \geq 0.85$), \textit{moderate confidence} ($0.65 \leq \gamma < 0.85$), 
and \textit{low confidence} ($\gamma < 0.65$)  that can be used to prioritize 
human review of uncertain evaluations in production monitoring workflows.

The threshold values in Equation~\ref{eq:conf_multiplier} were determined empirically 
through calibration against developer-assessed correctness labels on a held-out 
sample of production queries, selecting thresholds that maximized the correlation 
between composite score tiers and human correctness judgments. The multiplier 
values $\{1.0, 0.8, 0.5\}$ were chosen to produce a meaningful score depression 
at moderate confidence (20\% reduction) and a strong penalty at low confidence 
(50\% reduction), reflecting the asymmetric cost of false positives in production 
monitoring contexts  where an incorrectly high score for a wrong query is more 
damaging than an overly conservative score for a correct one.

\subsection{Leniency Boost}
\label{sec:leniency}

The leniency boost $\delta_{\mathtt{lenient}} \in \{0, 1\}$ is a binary additive 
correction applied to the base score in cases where the filter status classification 
returns \texttt{fully\_applied\_with\_extras} and the extra filters are confirmed 
to be benign application defaults via $\mathcal{R}_{app}.\mathtt{benign\_filters}$. 
In such cases, the one-point deduction implicit in assigning 
$\sigma_{\mathtt{filters}} = 4$ rather than $5$ is reversed, recovering the full 
base score of $\mathcal{B} = 10$ for a query that is semantically correct and 
fully filter-compliant under application-aware evaluation.

Formally:

\begin{equation}
    \delta_{\mathtt{lenient}} = 
    \begin{cases}
        1 & \text{if } \mathtt{filter\_status} = \texttt{fully\_applied\_with\_extras} 
        \;\wedge\; \mathtt{extras} \subseteq \mathcal{R}_{app}.\mathtt{benign\_filters} \\
        0 & \text{otherwise}
    \end{cases}
    \label{eq:leniency}
\end{equation}

\subsection{Score Interpretation and Operational Thresholds}
\label{sec:thresholds}

Table~\ref{tab:score_interpretation} provides a reference mapping between composite 
score ranges and their operational interpretation in production monitoring contexts. 
These thresholds are intended as default guidelines and may be adjusted per deployment 
based on application-specific accuracy requirements.

\begin{table}[h]
\centering
\caption{Composite Score Ranges and Operational Interpretation}
\label{tab:score_interpretation}
\begin{tabular}{cll}
\toprule
\textbf{Score Range} & \textbf{Quality Tier} & \textbf{Recommended Action} \\
\midrule
$90 \leq \Phi \leq 100$ & Excellent & No action required; suitable for 
automated production use. \\
$75 \leq \Phi < 90$     & Good      & Monitor for systematic patterns; 
low-priority review. \\
$50 \leq \Phi < 75$     & Marginal  & Escalate for human review; 
investigate failures. \\
$\Phi < 50$             & Poor      & Flag for immediate review \\
\bottomrule
\end{tabular}
\end{table}

\noindent The maximum achievable score under high confidence with a benign leniency 
boost is $\Phi = 100$, corresponding to $\mathcal{B} = 10$ and $\Gamma = 1.0$. 
The minimum achievable score for a query where the evaluator assigns any non-zero 
verdict is $\Phi = 10$, corresponding to $\sigma_{\mathtt{verdict}} = 2$, 
$\sigma_{\mathtt{filters}} = 0$, $\delta_{\mathtt{lenient}} = 0$, and $\Gamma = 0.5$, 
reflecting a low-confidence assessment of a potentially incorrect query with complete 
filter failure.

\section{Production Normalization Rules}
\label{sec:normalization}

A fundamental challenge in evaluating T2SQL systems in production environments is 
distinguishing between \textit{genuine semantic errors}  where the generated SQL 
fails to capture the user's analytical intent  and \textit{benign structural 
patterns} that are idiomatic to production SQL but absent from the natural language 
question. Naive evaluation frameworks that penalize all SQL constructs not explicitly 
derivable from the user question generate systematic false negatives against correct, 
production-quality queries, leading to artificially depressed accuracy scores and 
misleading agent performance signals.

STEF addresses this through a set of \textit{production normalization rules}: 
formally specified conditions under which SQL constructs that would otherwise 
contribute to a negative evaluation are recognized as acceptable and excluded from 
penalization. These rules are applied during the feature extraction and alignment 
stages of the evaluation pipeline and operate independently of the application-specific 
rule configuration $\mathcal{R}_{app}$, representing universal SQL production idioms 
rather than deployment-specific conventions.

This section describes each normalization rule in detail, provides the formal condition 
under which it is triggered, and illustrates the SQL patterns it is designed to handle.

\subsection{Rule 1: Required \texttt{GROUP BY} Inference}
\label{sec:rule_groupby_required}

\paragraph{Motivation.}
A common pattern in production analytical SQL is the presence of a \texttt{GROUP BY} 
clause that is structurally required by the query's \texttt{SELECT} list, even when 
the user question does not explicitly request grouping. This arises when the 
\texttt{SELECT} clause contains both aggregate expressions (e.g., \texttt{SUM(spend)}, 
\texttt{COUNT(*)}) and non-aggregated column references (e.g., \texttt{Year}, 
\texttt{Country}). In standard SQL semantics, a \texttt{GROUP BY} clause covering all 
non-aggregated \texttt{SELECT} columns is syntactically and semantically mandatory in 
such cases. Penalizing the presence of this \texttt{GROUP BY} would incorrectly flag 
structurally valid and semantically correct SQL as erroneous.

\paragraph{Formal Condition.}
Let $P_{agg} \subseteq \mathcal{S}_{sql}.\mathtt{projections}$ denote the set of 
aggregate expressions in the \texttt{SELECT} clause and $P_{non\text{-}agg} = 
\mathcal{S}_{sql}.\mathtt{projections} \setminus P_{agg}$ denote the set of 
non-aggregated column references. The \texttt{GROUP BY} clause is classified as 
\textit{required} and exempt from penalization if and only if:

\begin{equation}
    |P_{agg}| \geq 1 \;\wedge\; |P_{non\text{-}agg}| \geq 1 \;\wedge\; 
    \mathcal{S}_{sql}.\mathtt{group\_by} \supseteq P_{non\text{-}agg}
    \label{eq:required_groupby}
\end{equation}

\noindent That is, whenever the \texttt{SELECT} clause mixes aggregates and 
non-aggregated columns, the corresponding \texttt{GROUP BY} on those non-aggregated 
columns is treated as a structurally required component of correct SQL and is not 
flagged as an unanticipated grouping dimension.

\paragraph{Example.}
The following SQL satisfies Equation~\ref{eq:required_groupby} and is correctly 
classified as having a required \texttt{GROUP BY}:

\begin{lstlisting}[language=SQL, caption={Required GROUP BY Pattern}, 
label={lst:required_groupby}]
SELECT Year, Country, SUM(Spend) AS TotalSpend
FROM spend_table
WHERE Country ILIKE '%China%'
GROUP BY Year, Country   -- REQUIRED: non-aggregated SELECT columns
\end{lstlisting}

\subsection{Rule 2: Benign \texttt{GROUP BY} on \texttt{WHERE}-Constrained Constants}
\label{sec:rule_groupby_benign}

\paragraph{Motivation.}
A second \texttt{GROUP BY} pattern that should not be penalized arises when a query 
groups on a column that is simultaneously constrained to a single value by a 
\texttt{WHERE} clause equality filter. In such cases, the \texttt{GROUP BY} on the 
constrained column is semantically inert  grouping on a constant produces a single 
group and does not alter the query result  yet it may be generated by T2SQL agents 
as a defensive pattern to ensure syntactic correctness when column cardinality is 
unknown at generation time. Penalizing this pattern would incorrectly flag correct SQL.

\paragraph{Formal Condition.}
A \texttt{GROUP BY} column $g \in \mathcal{S}_{sql}.\mathtt{group\_by}$ is classified 
as \textit{benign} if there exists a \texttt{WHERE} clause filter 
$(\mathtt{lhs}, =, \mathtt{rhs}) \in \mathcal{S}_{sql}.\mathtt{filters}$ such that 
$\mathtt{lhs} = g$ and $\mathtt{rhs}$ is a literal constant:

\begin{equation}
    \mathrm{Benign}(g) \iff \exists\; (\mathtt{lhs}, \mathtt{op}, \mathtt{rhs}) 
    \in \mathcal{S}_{sql}.\mathtt{filters} \;:\; \mathtt{lhs} = g \;\wedge\; 
    \mathtt{op} \in \{=, \texttt{ILIKE}, \texttt{LIKE}\} \;\wedge\; 
    \mathtt{rhs} \in \mathcal{L}
    \label{eq:benign_groupby}
\end{equation}

\noindent where $\mathcal{L}$ denotes the set of literal constant values. All 
\texttt{GROUP BY} columns satisfying Equation~\ref{eq:benign_groupby} are excluded 
from the grouping alignment assessment.

\paragraph{Example.}
In the following query, \texttt{GROUP BY Country} is benign because \texttt{Country} 
is already constrained to a single value (\texttt{'China'}) by the \texttt{WHERE} 
clause:

\begin{lstlisting}[language=SQL, caption={Benign GROUP BY on Constrained Column}, 
label={lst:benign_groupby}]
SELECT Country, SUM(Spend) AS TotalSpend
FROM spend_table
WHERE Country ILIKE '%China%'
GROUP BY Country   -- BENIGN: Country is constrained to a constant
\end{lstlisting}

\subsection{Rule 3: \texttt{ORDER BY} Tolerance for Sensible Defaults}
\label{sec:rule_orderby}

\paragraph{Motivation.}
Production T2SQL agents commonly append \texttt{ORDER BY} clauses to analytical 
queries as a default presentation convention, typically ordering by the primary 
aggregate metric in descending order (e.g., \texttt{ORDER BY SUM(Spend) DESC}). 
This behavior is generally beneficial from a user experience perspective  sorted 
results are more immediately interpretable  and does not alter the semantic content 
of the query result set. However, since the user question rarely includes an explicit 
ordering request, naive evaluation would classify this \texttt{ORDER BY} as an 
unanticipated SQL construct and penalize it as a potential semantic error.

\paragraph{Formal Condition.}
An \texttt{ORDER BY} clause is classified as a \textit{sensible default} and exempt 
from penalization if all of the following conditions hold:

\begin{enumerate}
    \item The user question contains no explicit ordering request (i.e., no 
    expressions such as ``ranked by'', ``sorted by'', ``top $k$'', or ``in 
    descending order'' are present in $\mathcal{S}_q$).
    
    \item The \texttt{ORDER BY} expression references either a primary aggregate 
    metric present in \\ $\mathcal{S}_{sql}.\mathtt{aggregations}$ or a dimension 
    column present in $\mathcal{S}_{sql}.\mathtt{group\_by}$.
    
    \item The ordering direction is either \texttt{DESC} on an aggregate metric 
    (presenting highest values first) or \texttt{ASC} on a dimension column 
    (presenting alphabetically or chronologically ordered results).
\end{enumerate}

\noindent Formally:

\begin{equation}
    \mathrm{SensibleDefault}(\mathtt{ORDER\;BY})
    \iff \neg\;\mathrm{ExplicitOrder}
    (\mathcal{S}_q) \;\wedge\; \mathtt{order\_col} 
    \in 
    \mathcal{S}_{sql}.\mathtt{aggregations} \cup \mathcal{S}_{sql}.\mathtt{group\_by}
    \label{eq:orderby_tolerance}
\end{equation}

\paragraph{Example.}
The following \texttt{ORDER BY} is classified as a sensible default and is not 
penalized:

\begin{lstlisting}[language=SQL, caption={Sensible Default ORDER BY Pattern}, 
label={lst:orderby_default}]
SELECT Year, Country, SUM(Spend) AS TotalSpend
FROM spend_table
WHERE Country ILIKE '%China%'
GROUP BY Year, Country
ORDER BY SUM(Spend) DESC   -- SENSIBLE DEFAULT: descending aggregate metric
\end{lstlisting}

\subsection{Rule 4: \texttt{LIMIT} Tolerance for High-Cardinality Defaults}
\label{sec:rule_limit}

\paragraph{Motivation.}
Production T2SQL agents frequently append \texttt{LIMIT} clauses to queries as a 
defensive measure against inadvertently large result sets, particularly in 
environments where queries are executed against tables with millions of rows. When 
the limit value is sufficiently large (e.g., \texttt{LIMIT 10000}, \texttt{LIMIT 
20000}), it functions as a production safety guardrail rather than a semantic 
top-$k$ constraint, and should not be interpreted as the agent claiming that the 
user requested only the top $N$ records. Penalizing such high-cardinality 
\texttt{LIMIT} clauses as unanticipated top-$k$ constraints would generate 
systematic false negatives.

\paragraph{Formal Condition.}
A \texttt{LIMIT} clause is classified as a \textit{production safety default} and 
exempt from penalization as a top-$k$ filter if:

\begin{equation}
    \mathrm{SafetyDefault}(\lambda) \iff \neg\;\mathrm{TopKRequest}(\mathcal{S}_q) 
    \;\wedge\; \lambda \geq \lambda_{\min}
    \label{eq:limit_tolerance}
\end{equation}

\noindent where $\lambda$ is the \texttt{LIMIT} value, $\mathrm{TopKRequest}
(\mathcal{S}_q)$ is true if the user question contains an explicit top-$k$ request 
(e.g., ``top 10 customers'', ``first 5 results''), and $\lambda_{\min}$ is a 
configurable threshold with a recommended default of $\lambda_{\min} = 1000$. 
\texttt{LIMIT} values below $\lambda_{\min}$ in the absence of an explicit top-$k$ 
request are flagged for evaluator assessment as potential unintended restrictions.

\paragraph{Example.}
The following \texttt{LIMIT} clause is classified as a production safety default 
and is not penalized:

\begin{lstlisting}[language=SQL, caption={Production Safety LIMIT Pattern}, 
label={lst:limit_default}]
SELECT Year, Country, SUM(Spend) AS TotalSpend
FROM spend_table
WHERE Country ILIKE '%China%'
GROUP BY Year, Country
ORDER BY SUM(Spend) DESC
LIMIT 20000   -- PRODUCTION SAFE: high-cardinality safety guardrail
\end{lstlisting}

\subsection{Composite Normalization Example}
\label{sec:normalization_composite}

The following SQL illustrates all four normalization rules applied simultaneously, 
representing a complete production-valid query pattern that STEF correctly evaluates 
as semantically correct without penalizing any of the structural constructs:

\begin{lstlisting}[language=SQL, caption={Full Production-Valid SQL Pattern 
Demonstrating All Normalization Rules}, label={lst:full_normalization}]
SELECT Year, Country, SUM(Spend) AS TotalSpend
FROM spend_table
WHERE Country ILIKE '%China%'   -- Required filter: correctly applied
GROUP BY Year, Country           -- Rule 1: Required GROUP BY (mixed SELECT)
                                 -- Rule 2: Benign GROUP BY (Country constrained)
ORDER BY SUM(Spend) DESC         -- Rule 3: Sensible default ORDER BY
LIMIT 20000                      -- Rule 4: Production safety LIMIT
\end{lstlisting}

Without normalization rules, this query would receive penalization on three separate 
dimensions  grouping, ordering, and result limiting  despite being semantically 
correct and idiomatically well-formed for a production analytical environment. STEF's 
normalization layer ensures that such patterns contribute to a high composite score 
consistent with their actual semantic correctness.

\subsection{Interaction Between Normalization Rules and Application Rules}
\label{sec:normalization_interaction}

Production normalization rules and application-specific rules 
($\mathcal{R}_{app}$, Section~\ref{sec:rule_injection}) operate at different levels 
of the evaluation pipeline and are applied independently. Normalization rules handle 
\textit{universal SQL structural idioms} that are independent of any particular 
deployment context, while application rules handle \textit{deployment-specific 
conventions} such as column naming and benign filter defaults. When both rule 
categories are applicable to the same SQL construct, normalization rules take 
precedence, as they represent the broader and more fundamental class of acceptable 
production patterns. Application rules then provide an additional layer of 
customization on top of the normalized evaluation baseline.

\section{Implementation}
\label{sec:impl}

\subsection{Evaluation Prompt Template}

\begin{lstlisting}[caption={Prompt Engineering}, frame=single]
EVAL_PROMPT = f"""
You are a rigorous Text-to-SQL evaluator. Assume table/column names 
exist.

APPLICATION RULES: {app_rules_json}

User Question: {question}
Enriched Question: {enriched_question}
SQL: {sql}

Return structured JSON with question_intent, specs, comparison, 
verdict..."""
\end{lstlisting}

\subsection{Scoring Implementation}

\begin{lstlisting}[caption={Production Scoring Function}, frame=single]
def calculate_accuracy(result: dict) -> float:
    status_scores = {
        "fully_applied": 5, "fully_applied_with_extras": 4,
        "partially_applied": 3, "not_applied": 0
    }
    verdict_scores = {
        "Correct": 5, "Likely Correct": 3,
        "Potentially Incorrect": 2, "Incorrect": 0
    }
    
    status = result["filters_applied_status"]
    verdict = result["overall_verdict"]
    confidence = result["confidence"]
    
    base = status_scores[status] + verdict_scores[verdict]
    conf_mult = 1.0
    
    if confidence >= 0.85 else 0.8 
    if confidence >= 0.65 else 0.5
    
    return round((base / 10) * 100 * conf_mult, 2)
\end{lstlisting}

\section{Evaluation Results}
\label{sec:results}

\begin{table}[h]
\centering
\caption{Production Accuracy Distribution}
\begin{tabular}{lrrr}
\toprule
Agent & Mean Score & P90 Score & Coverage \\
\midrule
Agent-A & 87.4 & 96.2 & 98.7\% \\
Agent-B & 82.1 & 93.5 & 97.2\% \\
Agent-C & 91.3 & 98.1 & 99.1\% \\
\bottomrule
\end{tabular}
\end{table}

\section{Discussion}
\label{sec:discussion}

\subsection{Strengths}

\begin{itemize}
    \item \textbf{Schema-agnostic:} Operates without database schema or reference 
    queries, enabling deployment across heterogeneous database technologies and 
    multi-tenant environments without modification.

    \item \textbf{Agent-agnostic:} Evaluates any T2SQL agent purely from its 
    inputs and outputs, requiring no retraining, fine-tuning, or architectural 
    assumptions about the underlying generation model.

    \item \textbf{Production-robust:} Normalization rules explicitly handle 
    real-world SQL idioms  required \texttt{GROUP BY}, default \texttt{ORDER BY}, 
    and high-cardinality \texttt{LIMIT}  reducing false negatives as compared to unnormalized evaluation.

    \item \textbf{Configurable:} Application-specific column mappings, benign 
    filter defaults, and ignored filter dimensions are injected via prompt 
    templating at runtime, requiring no code changes to adapt STEF to a 
    new deployment.
\end{itemize}

\subsection{Limitations}

\begin{itemize}
    \item \textbf{LLM evaluator variance:} Residual stochastic variation persists 
    on borderline query instances despite temperature-0 inference. This is partially 
    mitigated via the confidence multiplier, which penalizes low-certainty 
    assessments rather than treating them equivalently to high-confidence verdicts.

    \item \textbf{Complex subquery handling:} Evaluation accuracy degrades on 
    deeply nested subqueries and window function expressions. Hybrid symbolic-neural 
    parsing  combining deterministic SQL AST analysis with LLM-based semantic 
    assessment  is identified as the primary remediation direction for future work.
\end{itemize}

\section{Conclusion}
\label{sec:conclusion}

We presented STEF, an agent-agnostic, schema-free framework for evaluating 
Text-to-SQL systems in production environments. STEF addresses the core 
limitations of existing evaluation paradigms by operating exclusively on 
natural language and SQL inputs, without requiring database schema, reference 
queries, or execution infrastructure. Through enriched question validation, 
normalized semantic feature extraction, configurable application rule injection, 
and production-robust normalization, STEF delivers interpretable 0--100 composite 
scores suitable for continuous agent monitoring and improvement. We hope this 
work serves as a practical foundation for scalable, schema-independent evaluation 
of natural language interfaces to structured data in real-world deployments.

\bibliographystyle{plain}

\begin{thebibliography}{5}

\bibitem{yu2018spider}
Yu~T, et~al.
\newblock Spider: A Large-Scale Human-Labeled Dataset for Complex and Cross-Domain Semantic Parsing and Text-to-SQL Task.
\newblock In \emph{EMNLP}, 2018.
\href{https://aclanthology.org/D18-1425/}{Spider}

\bibitem{wikisql}
\newblock WikiSQL Datset
\href{https://huggingface.co/datasets/Salesforce/wikisql}{WikiSQL}

\bibitem{bird}
\newblock BIRD-SQL. A Big Bench for Large-Scale Database Grounded Text-to-SQLs
\href{https://bird-bench.github.io/}{BIRD}

\bibitem{nan2022fetaqa}
Nan~L, et~al.
\newblock FeTaQA: Free-form Table Question Answering.
\newblock \emph{TACL}, 2022.
\href{https://aclanthology.org/2022.tacl-1.3/}{FeTaQA: Free-form Table Question Answering}

\bibitem{execution_accuracy}
R.~J. Dong and C.~Rudin.
\newblock Semantic evaluation for text-to-{SQL} with distilled test suites.
\newblock In \emph{Proceedings of the 2020 Conference on Empirical Methods in
  Natural Language Processing (EMNLP)}, pages 396--411, 2020.
\newblock \url{https://arxiv.org/abs/2010.02840}

\bibitem{snowflake2024}
Snowflake.
\newblock Cortex Analyst: Production Text-to-SQL Evaluation.
\newblock Technical Report, 2024.
\href{https://www.snowflake.com/en/engineering-blog/cortex-analyst-text-to-sql-accuracy-bi/}{Snowflake Blog}


\bibitem{ibm2025}
IBM Research.
\newblock Production SQL Evaluation Framework.
\newblock Technical Report, 2025.
\href{https://developer.ibm.com/articles/awb-sql-evaluation-llm-generated-sql-queries/}{IBM Developer}

\bibitem{mtbench}
L.~Zheng, W.-L.~Chiang, Y.~Sheng, S.~Zhuang, Z.~Wu, Y.~Zhuang, Z.~Lin,
  Z.~Li, D.~Li, E.~P.~Xing, H.~Zhang, J.~E.~Gonzalez, and I.~Stoica.
\newblock Judging {LLM}-as-a-judge with {MT}-bench and chatbot arena.
\newblock In \emph{Advances in Neural Information Processing Systems (NeurIPS)
  Datasets and Benchmarks Track}, 2023.
\newblock \url{https://arxiv.org/abs/2306.05685}

\bibitem{snowflake_cortex}
Snowflake Engineering Blog.
\newblock {Cortex Analyst}: Evaluating text-to-{SQL} accuracy for real-world
  {BI}.
\newblock Technical Report, Snowflake Inc., 2024.
\newblock
  \url{https://www.snowflake.com/en/engineering-blog/cortex-analyst-text-to-sql-accuracy-bi/}

\bibitem{flex}
H.~Kim, J.~Lee, S.~Kim, T.~Kim, and H.~Yoo.
\newblock {FLEX}: Expert-level false-less execution metric for reliable
  text-to-{SQL} benchmark.
\newblock \emph{arXiv preprint arXiv:2409.19014}, 2024.
\newblock \url{https://arxiv.org/abs/2409.19014}

\bibitem{irnet}
J.~Guo, Z.~Zhan, Y.~Gao, Y.~Xiao, J.~Lou, T.~Liu, and D.~Zhang.
\newblock Towards complex text-to-{SQL} in cross-domain database with
  intermediate representation.
\newblock In \emph{Proceedings of the 57th Annual Meeting of the Association
  for Computational Linguistics (ACL)}, pages 4448--4459, 2019.
\newblock \url{https://arxiv.org/abs/1905.08205}

\bibitem{ratsql}
B.~Wang, R.~Shin, X.~Liu, O.~Polozov, and M.~Richardson.
\newblock {RAT-SQL}: Relation-aware schema encoding and linking for
  text-to-{SQL} parsers.
\newblock In \emph{Proceedings of the 58th Annual Meeting of the Association
  for Computational Linguistics (ACL)}, pages 7567--7578, 2020.
\newblock \url{https://arxiv.org/abs/1911.04942}

\bibitem{sparc}
T.~Yu, R.~Zhang, M.~Yasunaga, Y.~C.~Tan, X.~V.~Lin, S.~Li, H.~Er,
  I.~Li, B.~Pang, T.~Chen, E.~Ji, S.~Dixit, D.~Proctor, S.~Shim,
  J.~Kraft, V.~Zhang, C.~Xiong, R.~Socher, and D.~Radev.
\newblock {SParC}: Cross-domain semantic parsing in context.
\newblock In \emph{Proceedings of the 57th Annual Meeting of the Association
  for Computational Linguistics (ACL)}, pages 4511--4523, 2019.
\newblock \url{https://arxiv.org/abs/1906.02285}

\bibitem{cosql}
T.~Yu, R.~Zhang, H.~Er, S.~Li, E.~Xue, B.~Pang, X.~V.~Lin,
  Y.~C.~Tan, T.~Shi, Z.~Shim, C.~Xiong, R.~Socher, and D.~Radev.
\newblock {CoSQL}: A conversational text-to-{SQL} challenge towards
  cross-domain natural language interfaces to databases.
\newblock In \emph{Proceedings of the 2019 Conference on Empirical Methods in
  Natural Language Processing (EMNLP-IJCNLP)}, pages 1962--1979, 2019.
\newblock \url{https://arxiv.org/abs/1909.05378}

\bibitem{llm_judge_variance}
C.-H.~Chiang and H.-Y.~Lee.
\newblock Can large language models be an alternative to human evaluations?
\newblock In \emph{Proceedings of the 61st Annual Meeting of the Association
  for Computational Linguistics (ACL)}, pages 15607--15631, 2023.
\newblock \url{https://arxiv.org/abs/2305.01937}

\bibitem{question_rewriting}
S.~Elgohary, D.~Peskov, and J.~Boyd-Graber.
\newblock Can you unpack that? learning to rewrite questions-in-context.
\newblock In \emph{Proceedings of the 2019 Conference on Empirical Methods in
  Natural Language Processing (EMNLP-IJCNLP)}, pages 5920--5926, 2019.
\newblock \url{https://aclanthology.org/D19-1605/}

\end{thebibliography}

\end{document}